\documentclass[a4paper,conference]{IEEEtran}

\IEEEoverridecommandlockouts
\usepackage[noadjust]{cite}
\usepackage{amsmath,amssymb,amsfonts}

\usepackage{algorithm,algorithmic}

\def\BibTeX{{\rm B\kern-.05em{\sc i\kern-.025em b}\kern-.08em
    T\kern-.1667em\lower.7ex\hbox{E}\kern-.125emX}}

\usepackage{textcomp}
\usepackage{xcolor}
\usepackage{url}
\usepackage{microtype}
\usepackage{graphicx}
\usepackage{subfigure}

\usepackage{booktabs} % for professional tables
\usepackage{amsmath}
\usepackage{amsfonts}
\usepackage{tikz}
\usetikzlibrary{angles,quotes}

\usepackage{blindtext}
\usepackage[inline]{enumitem}
\usepackage{color}
\usepackage{xcolor,colortbl}
\newcommand{\bz}{{\mathbf z}}
\newcommand{\bx}{{\mathbf x}}
\newcommand{\comment}[1]{{}}

\def\BibTeX{{\rm B\kern-.05em{\sc i\kern-.025em b}\kern-.08em
    T\kern-.1667em\lower.7ex\hbox{E}\kern-.125emX}}

\usepackage{fancyhdr}
\fancypagestyle{firstpage}{
\lhead{25th International Conference on Pattern Recognition, ICPR 2020}
\rhead{}
}

\begin{document}

\title{Generative Latent Implicit Conditional Optimization when Learning from Small Sample}

\author{\IEEEauthorblockN{Idan Azuri}
	\IEEEauthorblockA{\textit{School of Computer Science and Engineering,} \\
		\textit{The Hebrew University,}	Jerusalem, Israel \\
	idan.azuri@cs.huji.ac.il}
	\and
	\IEEEauthorblockN{Daphna Weinshall}
	\IEEEauthorblockA{\textit{School of Computer Science and Engineering,} \\
		\textit{The Hebrew University,}
		Jerusalem, Israel \\
	daphna@cs.huji.ac.il}
}

\maketitle

\begin{abstract}
    	We revisit the long-standing problem of \textit{learning from small sample}, to which end we propose a novel method called \textit{GLICO} (Generative Latent Implicit Conditional Optimization). \textit{GLICO} learns a mapping from the training examples to a latent space, and a generator that generates images from vectors in the latent space. Unlike most recent works, which rely on access to large amounts of unlabeled data, \textit{GLICO} does not require access to any additional data other than the small set of labeled points.	In fact, \textit{GLICO} learns to synthesize completely new samples for every class using as little as 5 or 10 examples per class, with as few as 10 such classes without imposing any prior. \textit{GLICO} is then used to augment the small training set while training a classifier on the small sample. To this end our proposed method samples the learned latent space using spherical interpolation, and generates new examples using the trained generator. Empirical results show that the new sampled set is diverse enough, leading to improvement in image classification in comparison with the state of the art, when trained on small samples obtained from CIFAR-10, CIFAR-100, and CUB-200.
			
\end{abstract}

\maketitle
\thispagestyle{firstpage}

%\begin{IEEEkeywords}
%	small sample, generative models, latent optimization, classification, data augmentation
%\end{IEEEkeywords}

\section{Introduction}
Modern deep Convolutional Neural Networks (CNNs) define the state of the art in image classification, as well as many other problems in a wide range of applications. Typically enormous amounts of labeled data are used to train the networks. It is not obvious whether this success can be replicated in domains where the resource of labeled data is not widely available. While hardly unexplored, the question of learning from a small sample remains a very important and challenging problem, not least so in the context of deep learning and image classification. The question remains relevant to date, as collecting a very large set of images may be difficult due to issues of privacy, image quality, geographical location, time period, or copyright status. The difficulties are further aggravated in applications that require the acquisition and processing of a new custom dataset of images, which may be a highly costly task. 

In the strict small sample settings, the learner has access to a small number of labeled examples from each class, possibly as few as 5 or 10, and the number of classes can also be rather small. This setting is substantially different from the two related settings of semi-supervised learning and the few-shots learning scenario. In the few-shot scenario, the learner has access to a large number of labeled examples from classes not participating in the current classification task, while in the semi-supervised scenario the learner typically has access to a large number of unlabeled examples. Thus most few-shot algorithms rely on transfer learning from tens of thousands of labeled training examples, while most semi-supervised algorithms transfer knowledge from the distribution of the unlabeled data.  

Methods designed to address the strict small sample scenario cannot rely on transfer learning from large amounts of peripheral data. Different approaches have been explored to address this problem, as reviewed in the next section. The problem can be notably alleviated by imposing a strong prior on the model. Unfortunately, in many applications we face an unknown domain and such prior knowledge is not available. 

\begin{figure}[t]
		\centering
		\includegraphics[width=0.75\linewidth]{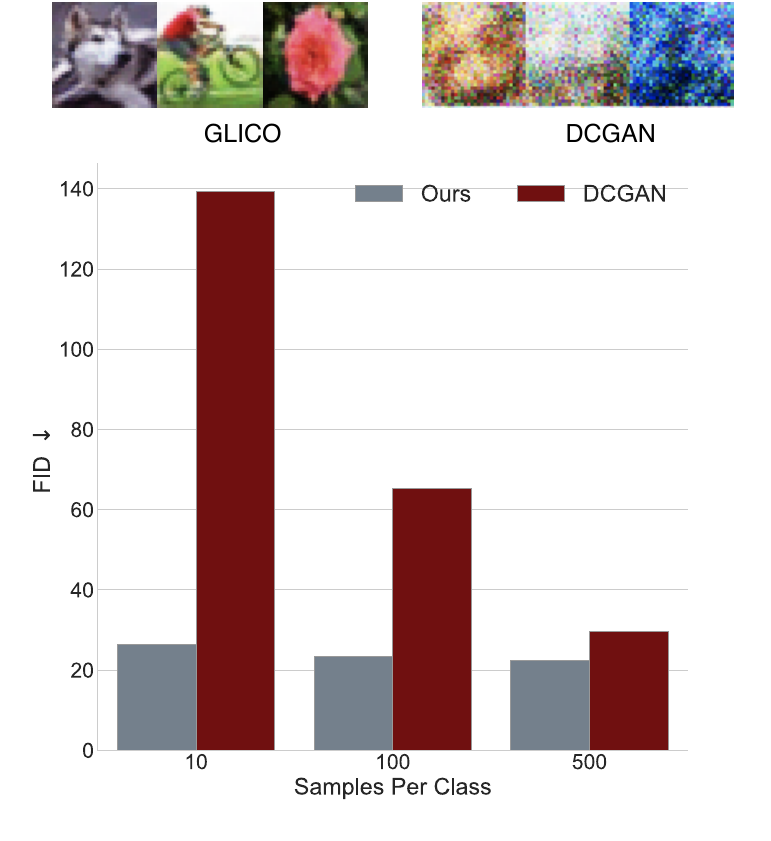}

% 	\vskip -0.15in
	\caption{The quality of image generation in a small sample settings. We compare our method to DCGAN with the same generator. The quality of the generated images is measured using the FID score \cite{DBLP:journals/corr/HeuselRUNKH17} (lower is better). Top: new synthesized examples using our method (GLICO) and a widely used GAN model (DCGAN), both trained on CIFAR-100 with only 10 samples per class. Bottom: FID scores for each generator, when using 10, 100 and 500 examples from each class in CIFAR-100.}
	
	\label{fig: fid}
	\vskip -0.2in
\end{figure}

Data augmentation may help, reflecting the availability of some weak prior knowledge about the data. Methods employing semi-supervised \cite{odena2016semi} and transductive learning \cite{vapnik1998statistical} make use of unlabeled data, when available. Self-training approaches can also boost performance when labels are scarce. Finally, one may compute a generative model from the training data, and use it to generate new samples. This is the approach we explore in this study. The different approaches are not mutually exclusive and can be used in conjunction to further boost performance as shown in Section~\ref{sec: autoaug}.

\textbf{Why not GAN?}
Using generative models to augment a small training sample is very appealing, especially at present when very powerful deep generative models are becoming available. The problem is that in general, these models require a very large (possibly unlabeled) sample to achieve effective training, and therefore can only be used to augment the training set in the semi-supervised scenario. These models perform poorly in the strict small sample scenario, as can be seen in Fig.~\ref{fig: fid}. To obviate this problem, our method optimizes the latent space directly, similarly to \cite{pmlr-v80-bojanowski18a}. It is thus able to effectively synthesize quality new images in the small sample domain, outperforming the conditional DCGAN as can be seen in Fig.~\ref{fig: fid}, where image quality is measured using the FID score.

More specifically, our proposed method (see Fig.~\ref{fig:our_method}) learns a latent space code for each data point separately. To improve the properties of the learned latent space, we seek to push intra-class examples to lie close to each other and separated from inter-class examples as much as possible. This is accomplished by adding a classifier to the basic architecture as described in  Section~\ref{sec:glo}. The classifier is trained using the known multi-class labels of the data. We note that training is not adversarial, and therefore this classifier is not equivalent to the discriminator in the GAN (Generative Adversarial Network) architecture. %Additional modifications include the concatenation of noise to the latent space vector. 
The full model is described in Section~\ref{sec:model}. 

The architecture of the model reveals geometric properties that allow us to effectively sample specific areas in the latent space and synthesize novel images as explained in Section~\ref{sec:glo}. 
%[not sure if it is relevant] In this way our method bears some similarity to \cite{chawla2002smote}, a method designed to deal with the problem of imbalanced datasets by creating synthetic minority class examples.
The empirical evaluation described in Section~\ref{sec:evalution} shows the success of our model in improving classification performance while training with a small sample, especially in the extreme conditions where the sample size is very small indeed.

The rest of the paper is organized as follows. In Section~\ref{sec:model} we describe \textit{GLICO}, a novel method for data synthesis which can be effectively trained from a few labeled points from each class. %Our method can make use of unlabeled data when available, further benefiting from additional data in the semi-supervised and transductive learning scenarios. 
In Section~\ref{sec:evalution} we describe how synthetic images are used to boost the training of a discriminative deep classifier. In Section~\ref{sec:results} we demonstrate the superior performance of our method when compared to alternative methods under extreme low sample conditions in the strict small sample scenario with no additional unlabeled data. We provide an ablation study and further investigate alternative design choices and their effect on classification performance. 
In Section~\ref{sec: autoaug} we show empirically that the synthetic examples produced by \textit{GLICO} are unlike those generated by conventional augmentation methods. Lastly, we investigate the contribution of adding unlabelled data to \textit{GLICO} in the semi supervised settings in Section~\ref{sec:other methods}.

% \paragraph{Our Contribution is the following:}
% \begin{enumerate*}[label={\roman*)},font={\color{black}\bfseries}]  
% 	\item We propose a novel data agnostic method for data generation in the low data regime. 
% 	\item We analyze the latent space of the proposed model \textit{GLICO}; we show how different design choices of our model affect the classification performance.
% 	\item We explore how the latent space of a generative model is influenced by transductive learning.
% 	\item We measure the effectiveness of adding a classifier to the latent optimization training process as a regularization method. 
% \end{enumerate*}

\section{Related Work}

Earlier work addressed the small sample scenario mainly in the context of Bayesian inference, see for example \cite{Baxter97abayesian, Thrun1998LearningTL, LiFergusPerona04}. %Bayesian methods rely on the existence of a large dataset which is correlated with the test set. Under this assumption, a prior can be established and used to identify test examples from unseen classes. 
These methods have been revisited in recent years in the context of \emph{few-shot learning}, where transfer learning can be exploited. In a commonly used setup of \emph{few-shot learning}, at train time there are many classes with many labeled examples from each class. At test time, some novel classes (usually 5) with only a few examples per class are given, alongside query images from the same novel classes. 
 
\textbf{Data Augmentation with Geometrical Transformations.} Focusing our attention back at the small sample problem, \emph{data augmentation} as discussed in \cite{Tanner1987TheCO} can be effectively used. %This include augmentation in the image domain for image classification \cite{Baird1993DocumentID, Krizhevsky2012ImageNetCW, Cubuk2018AutoAugmentLA}, time-series transformations for natural language processing tasks \cite{kobayashi-2018-contextual, charcter}, speech recognition \cite{park2019specaugment}, and machine translation \cite{fadaee2017data,wang-etal-2018-switchout}. %In the following literature review, we divide the prior work into two complementary directions: augmentation based on assumed invariance in the data, and the generation of a new sample using generative models. Metric learning can also be used for this purpose e.g., \cite{hertz2006learning,li2019revisiting}, where recently \cite{cosinesmalldata} showed that using the cosine distance may improve learning from small sample.
%
%\paragraph{Invariance-based augmentation.}
Common augmentation techniques for image classification include translating the image by a few pixels, adding Gaussian noise, and flipping the image horizontally or vertically \cite{Baird1993DocumentID, Krizhevsky2012ImageNetCW, Cubuk2018AutoAugmentLA}. %These techniques require prior knowledge of the data. For instance, vertical flip augmentation of the digit 6 will change its semantic meaning to 9. 
%Methods such as AutoAugment \cite{Cubuk2018AutoAugmentLA}, as well as Fast Augment \cite{Lim2019FastA,hataya2019faster} and Smart Augmentations \cite{lemley2017smart}, achieve further improvement by searching for the optimal set of augmentations in a predefined set of classical transformations.%, including random crop, flip, rotation, scale, and translation. The major drawback of such methods lies in the enormous space that needs to be searched, and the huge resources that are needed to perform an effective search. 
%
Additional image augmentation methods include \emph{Cutout} \cite{devries2017improved}, which randomly masks a square region in an image at every training step and thus affects the nature of the learned features, and \emph{Random Erasing} \cite{randomerasing}, where similarly to dropout randomly chosen rectangular regions in the image have their pixels erased or replaced by random values. \emph{MixUp} \cite{zhang2017mixup} uses Alpha-blending of two images to form a new image while regularizing the CNN to favor a simple linear behavior in between training images. \emph{MixMatch} \cite{berthelot2019mixmatch} augments MixUp by self training, generating “guessed labels” for each unlabeled example. We note that these methods are not always effective on very small datasets, and may even degrade classification performance as shown in Table~\ref{tab: multi-shot-cifar}.

\textbf{Data Augmentation with Generative Models.}
As alluded to above, generative models can be a powerful tool for data augmentation by making it possible under ideal conditions to sample new examples, see for example \cite{schwartz2018delta,antoniou2018augmenting,zhang2019dada,devries2017dataset,liu2018data,zhang2017mixup}.
%. Most of the methods which rely on generative models target the few-shot learning scenario. Among these methods, Delta-Encoder \cite{schwartz2018delta} investigates the use of a modified AutoEncoder: the model gets two examples from a known class and then learns the 'delta' between the two examples to generate new examples for novel classes. DAGAN \cite{antoniou2018augmenting} employs a GAN composed of a U-net generator and a DenseNet discriminator. \cite{zhang2019dada} modified the discriminator of a semi-supervised GAN to return $2N$ outputs, where the first $N$ elements denote class probability, and the remaining $N$ outputs denote the probability that the example is fake. Methods such as \cite{devries2017dataset,liu2018data} also make use of Auto-Encoders. When using an Adversarial AutoEncoder (AAE), one can populate "dead zones" in the latent space by initializing it uniformly and then applying MixUp \cite{zhang2017mixup}. 
These methods typically require large amounts of (labeled or unlabeled) data to be effective. 
They try to leverage generative models by sampling new synthetic examples from the learned distribution of the data. 
%However, the estimation of the latent space for a given dataset is challenging and requires many training examples. In the low-shot case, the estimated data distribution is less reliable, and it is, therefore, necessary to rely on the joint distribution of related classes. Thus one approach lets the learning model share its parameters across all classes, using the same generator in a GAN architecture, or the same encoder and decoder in an Auto-Encoder. 
Unfortunately in the small sample case, when transfer learning is not an option, the few labeled examples do not represent the true data distribution very reliably, resulting in poor generalization and low-quality synthetic data. Finally, the notorious instability of their training process and the heavy computational load make GANs less appealing for mere data augmentation.%, and the dependency on very large datasets \cite{Goodfellow2017NIPS2T}.

\section{Our Model}
\label{sec:model}

% *a high-level explanation of our method*
We propose a method for multi-class image classification from small sample, which consists of data augmentation with a generative model. Specifically, to use modern deep learning classifiers, which typically require large amounts of training data, we augment the small training set by sampling from a generative model. Our generative model is a latent optimization method that combines an auxiliary task of classification, and which can be trained effectively from a small sample. The architecture is designed so that it can benefit from both labeled and unlabeled data. With access to only small amounts of unlabeled data or none at all, our results surpass the state of the art. 
The code is available online\footnote{\url{https://github.com/IdanAzuri/glico-learning-small-sample}}.
% The code is available online\footnote{\url{https://anonymous.4open.science/r/7f821a4e-caee-4c91-b047-6d389fce1966/}}.	

\subsection{Model Overview}

\begin{figure}[t]
% 	\vskip -0.2in
	\begin{center}
		\centerline{\includegraphics[width=\columnwidth]{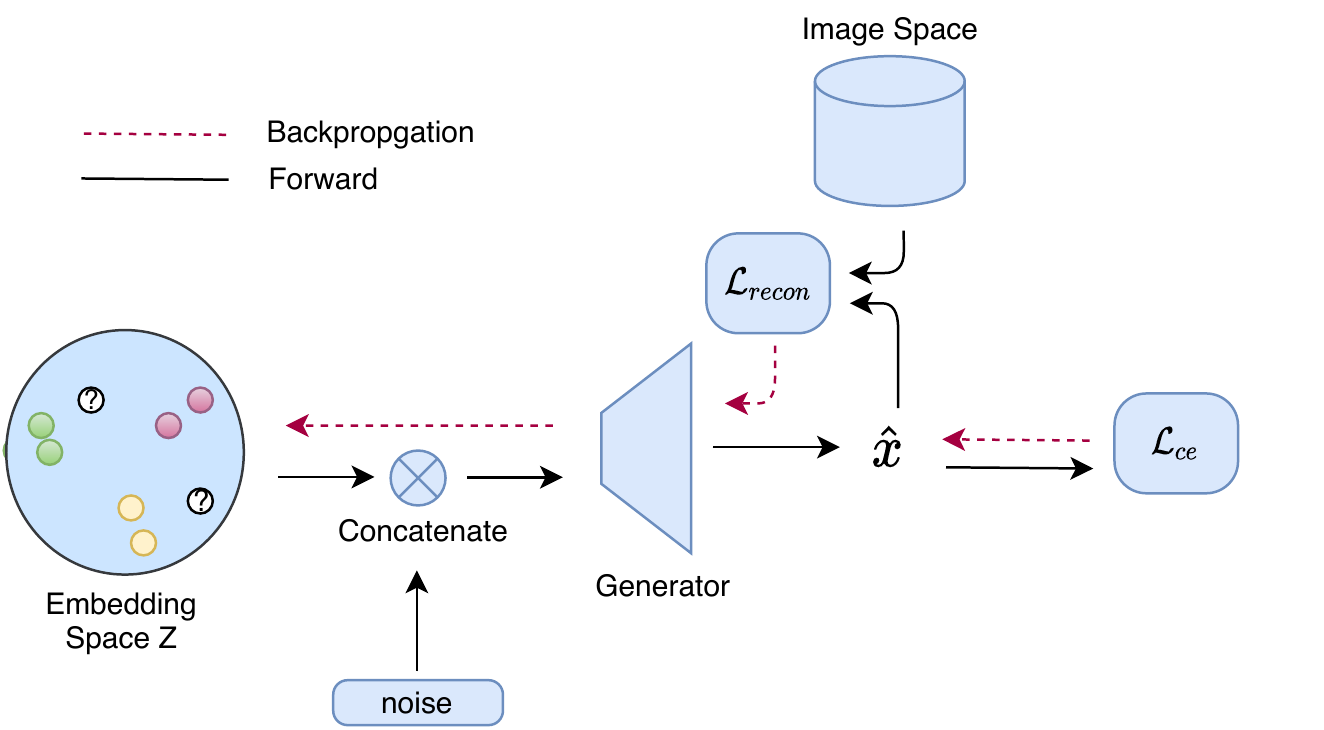}}
		\caption{Schematic illustration of our method. Every $\bz_i\in Z$, where $Z$ denotes the unit sphere, is mapped to a specific image $x_i$ in the image space. In this illustration, the color of each image frame marks the class label of the image, while black frames mark unlabeled images. The classifier is used to propagate error only when a label is given.}
		\label{fig:our_method}
	\end{center}
	\vskip -0.1in
\end{figure}

The generative model we use to sample new data includes a basic generative latent optimization architecture with generator $G_\theta$, and an added small classifier $f_\phi$ which is trained to classify the labeled data (see Fig.~\ref{fig:our_method}). Training is not adversarial, and therefore this classifier is nothing like the discriminator in the GAN architecture. The latent space is initialized randomly $\{\bz_i\in Z\}$ where $Z$ denotes the unit sphere in $\mathbb{R}^d$. Every vector $\bz_i$ is mapped to an image $\{ x_i\in X | x_i \in \mathbb{R}^{3 \times H \times W} \}$ from the given (small) training set.
		
The training process has two modes: With unlabeled data, the reconstruction loss in (\ref{eq:recon}) is used to train $G_\theta$ as done in \cite{pmlr-v80-bojanowski18a}. With labeled data, the reconstruction loss is augmented with the cross-entropy loss corresponding to the loss of the added classifier $f_\phi$ (see Fig.~\ref{fig:our_method}). 

Here we use the perceptual loss \cite{johnson2016perceptual} to measure the reconstruction loss. More specifically, in order to compute the perceptual loss we extract the activation vectors in layers ‘$conv_{1\_2}$’, ‘$conv_{2\_2}$’, ‘$conv_{3\_2}$’, ‘$conv_{4\_2}$’ and ‘$conv_{5\_2}$’ of a VGG-16 network. Denoting the output tensor of layer ‘$conv_{j\_2}$’ for input image x by $\xi_j(x)$, we compute the difference between the original image and its reconstructed version by:
\begin{equation}
	\label{eq: percep}
	\mathcal{L}_{\textit{percep}}(x_i,\bz_i;\theta) = \sum_j \lambda_j ||\xi_j(G_\theta([\bz_i, \varepsilon]))-\xi_j(x_i))||_1  
\end{equation}
Above $\theta$ denotes the parameters of the generator $G_\theta$, and $\lambda_j$ the weight of layer $j$ (usually the weighted average).

Our method is described in Alg.~\ref{alg:our_method}. Its components are described next.
\subsection{Generative Latent Optimization}
\label{sec:glo}

\begin{algorithm}[bt]
	\caption{GLICO. The algorithm learns codes $\{\bz_i\}_{i=1}^n$ by minimizing the reconstruction loss $\mathcal{L}_{\textit{percep}}$ of generator $G_\theta$, and the cross entropy loss $\mathcal{L}_{\textit{ce}}$ of discriminator $f_\phi$.}
	\label{alg:our_method}
	\begin{algorithmic}
		\STATE {\bfseries Input:} unlabeled data $P_{D_u}$, labeled data $P_{D_l}$, $\gamma$, epochs
		\STATE $n = |P_{D_l}|+ |P_{D_u}|$
		\STATE epoch = 0
		\STATE Initialize $\{\bz_i\}_{i=1}^n$ where $\{\bz_i\in Z: ||\bz_i||_2 =1 \}$
		\REPEAT
														        
		\FOR{$(x_i,y_i)\in P_{D_l}$}
		\STATE $\varepsilon \sim N(0,\sigma I)$
		\STATE $\mathcal{L}_{\textit{percep}} =\sum_j \lambda_j ||\xi_j(G_\theta([z_i,\varepsilon]))-\xi_j(x_i))||_1$
		\STATE $\mathcal{L}_{\textit{ce}} = \mathcal{L}_{CE}(f_{\phi}(G_\theta([z_i,\varepsilon])),y_i)$
		\STATE $\mathcal{L}= \mathcal{L}_{\textit{percep}} + \gamma \mathcal{L}_{\textit{ce}} $
		\STATE \text{Update} $\{\bz_i\}, \theta, \phi$
		\text{using the gradient of} $\mathcal{L}$
		\STATE   $\forall \bz_i \in Z$, $\bz_i:= \frac{\bz_i}{||\bz_i||_2} $
		\ENDFOR
		\STATE // Optional: transductive learning mode
		\FOR{$x_i\in P_{D_u}$}
		\STATE $\mathcal{L} =\sum_j \lambda_j ||\xi_j(G_\theta([z_i,\varepsilon]))-\xi_j(x_i))||_1$
		\STATE \text{Update} $\{\bz_i\}, \theta$  \text{using the gradient of} $\mathcal{L}$
		\STATE   $\forall \bz_i \in Z$, $\bz_i:= \frac{\bz_i}{||\bz_i||_2} $
		\ENDFOR
		\STATE $\textit{epoch} += 1$
		\UNTIL{$epoch > epochs$}
	\end{algorithmic}
\end{algorithm}

\textbf{Generative model.}
Generative Latent Optimization (GLO) 
is an effective and thin method for image reconstruction, relying on a small number of parameters. GLO maps every image $x_i$ from the dataset to a low-dimensional random vector $\bz_i$ in the latent space $Z$. It then passes the random vector into a generator $G_\theta( \cdot)$, which is optimized to minimize the reconstruction loss between $G_\theta(\bz_i)$ and $x_i$. 

Formally, let $\{x_1,x_2 \ldots x_n \} \in X$ denote a set of images where $x_i \in \mathbb{R}^{3 \times W \times H}$. Choose $n$ $d$-dimensional random vectors on the unit sphere $\{\bz_1,\bz_2, \ldots ,\bz_n \}\in Z$ where $Z  \subseteq \mathbb{R}^d$. Pair every image $x_i \in X$ with a random vector $\bz_i \in Z$, to achieve the mapping $\{(x_1,\bz_1), \dots, (x_n,\bz_n)\}$. Finally, learn jointly the parameters $\theta$ of the generator $G_\theta:Z \xrightarrow{} X$, where the optimal set $\{\bz_i\}$ and parameters $\theta$ are obtained by minimizing the following objective:
% \begin{equation}
% 	\begin{split}
% 		\label{eq:recon}
% 		\min_{\theta} \sum_{i=1}^n &\left [ \min_{\bz_i\in Z} \frac{1}{CHW}||x_i - G_{\theta} (\bz_i)||_p \right ] \\
% 		& \quad\quad s.t. \quad ||\bz_i||_2=1
% 	\end{split}
% \end{equation}
% \begin{equation}
% 	\begin{split}
% 		\label{eq:recon}
% 		\min_{\theta} \mathop{\mathbb{E}}_{i\in [n]} &\left [ \min_{\bz_i\in Z} ||x_i - G_{\theta} (\bz_i)||_p \right ] \\
% 		& \quad\quad s.t. \quad ||\bz_i||_2=1
% 	\end{split}
% \end{equation}
\begin{equation}
	\begin{split}
		\label{eq:recon}
		\min_{\theta} \sum_{i=1}^n &\left [ \min_{\bz_i\in Z} \mathcal{L}_{\textit{percep}}(x_i,\bz_i;\theta) \right ] \\
		& \quad\quad s.t. \quad ||\bz_i||_2=1
	\end{split}
\end{equation}
% The loss $\mathcal{L}_{\textit{percep}}(x_i,\bz_i;\theta)$, defined in (\ref{percep}), measures the reconstruction loss between $G_\theta(\bz_i)$ and $x_i$. We note that while in the original GLO the Laplacian pyramid loss is used instead of $\mathcal{L}_{\textit{percep}}$, the minimization of  $\mathcal{L}_{\textit{percep}}$ appears to yield more realistic results \cite{glann}.
% 		The loss $\mathcal{L}_{\textit{percep}}(x_i,\bz_i;\theta)$, defined in (\ref{percep}), measures the reconstruction loss between $G_\theta(\bz_i)$ and $x_i$.
While in GLO the Laplacian pyramid is used for reconstruction loss, we note that the minimization of $\mathcal{L}_{\textit{percep}}$ appears to yield more realistic results \cite{glann}, as compared to other image metrics in common use.
		
\begin{figure}[htbp]
	% 	\vskip -0.2in
	\begin{center}
		\centerline{\includegraphics[width=\columnwidth]{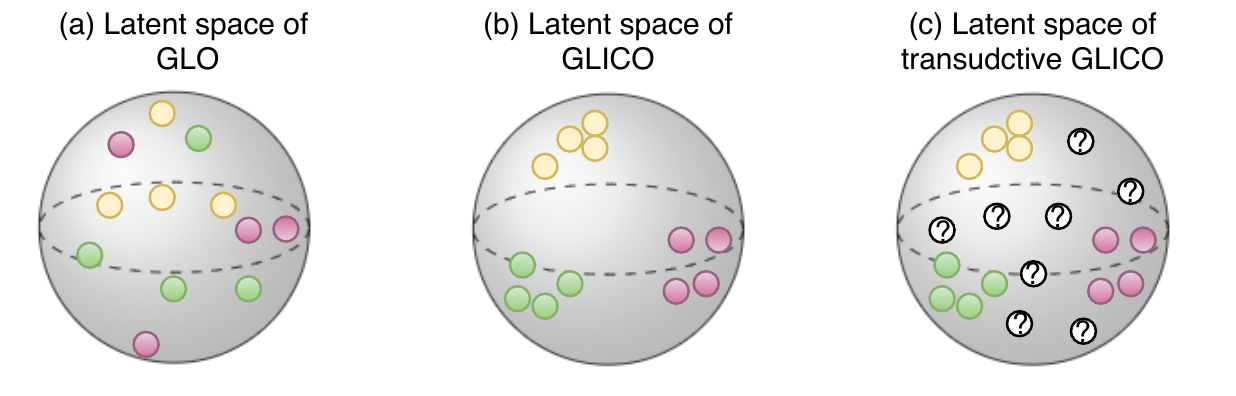}}
		\caption{Illustration of the latent space $Z$. (a) GLO: vectors $\bz_i\in Z$ do not have semantic meaning in $Z$. (b) Our method: vectors from the same class are grouped. (c) Our method in transductive mode. Notations: filled colored circles represent different labeled datapoints, where color corresponds to class identity. Black circles with the symbol "?" represent unlabeled datapoints.}
		\label{fig:embed}
	\end{center}
	\vskip -0.2in
\end{figure}

\textbf{Adding a Classifier.}
Possibly the main weakness of using GLO as a generative model is the relatively low quality of images generated when sampling new points in the latent space $Z$. The problem lies in the sparsity of the learned set $\{\bz_i\}_{i=1}^n$, which lacks structure since each $\bz_i$ is trained independently. To decrease the intra-class distances and increase the inter-class distances in the latent space representation, we propose the conditional model termed \textit{GLICO}. In this model, the generator $G_\theta$ is augmented by a weak classifier $f_\phi$, which is trained to classify the labeled data. When the label of $x_i$ is known, $\mathcal{L}_{\textit{percep}}$ in (\ref{eq:recon}) is replaced by $\mathcal{L}_{\textit{percep}}+\mathcal{L}_{\textit{ce}}$, where $\mathcal{L}_{\textit{ce}}$ is the cross-entropy loss of classifier $f_\phi$. %This augmentation is likely to increase the latent space density of similar label points around each $\bz_i$.
Fig.~\ref{fig:embed} illustrates the structure of the latent space when the different loss functions are used.

\textbf{Sampling the Latent Space.} 
Generating images based on randomly sampling the latent space, even when restricted to the immediate vicinity of $\{\bz_i\}_{i=1}^n$, produces low quality somewhat meaningless images. Therefore, instead of randomly sampling $Z$, we generate new image codes by interpolating between the known latent vectors $\{\bz_i\}_{i=1}^n$. Since the latent space $Z$ is a hyper-sphere, we employ to this end \textit{spherical linear interpolation} (\emph{slerp}) \cite{Shoemake:1985:ARQ:325165.325242}, which is defined as follows: 
\begin{equation}
	\label{eq:slerp}
	slerp(q1, q2;t) = q_1\frac{\sin{(1-t)\vartheta}}{\sin{\vartheta}}  + 
	q_2\frac{\sin{t\vartheta}}{\sin{\vartheta}}
\end{equation}
Above $t \in [0,1]$, and $\vartheta$ is the angle between $q_1$ and $q_2$, computed as $\vartheta = \cos^{-1} (q_1 \cdot q_2)$.

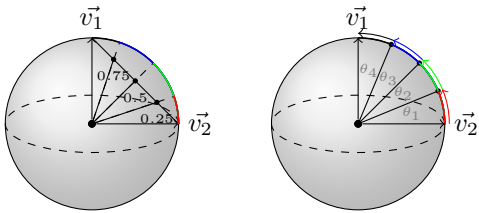
\begin{figure}[htbp]
	\centering
	\begin{tikzpicture}[scale=0.57]
		\shade[ball color = gray!40, opacity = 0.4] (0,0) circle (2cm);
		% Define radius
		\def\r{2}
		% Bloch vector
		%sep=point width
		\draw (0,0) node[circle,fill,inner sep=1] (orig) {} -- (\r/2,\r/2) node[circle,fill,inner sep=0.7    ,label=below:{\tiny$0.5$}] (a) {};
		\draw (0,0) node[circle,fill,inner sep=1] (orig) {} -- (1.5,0.5) node[circle,fill,inner sep=0.7,label=below:{\tiny$0.25$}] (b) {};
		\draw (0,0) node[circle,fill,inner sep=1] (orig) {} -- (0.5,1.5) node[circle,fill,inner sep=0.7,label=below:{\tiny$0.75$}] (c) {};

		\draw[dashed] (orig) -- (1.42, 1.42);
		\draw[dashed] (orig) -- (0.64, 1.90);
		\draw[dashed] (orig) -- (1.90, 0.64);
		% Sphere
		\draw (orig) circle (\r);
		\draw[dashed] (orig) ellipse (\r{} and \r/3);
														        
		% Axes
		% \draw[->] (orig) -- ++(-\r/5,-\r/3) node[below] (x1) {$x_1$};
		\draw[->] (orig) -- ++(\r,0) node[right] (v2) {$\vec{v_2}$};
		\draw[->] (orig) -- ++(0,\r) node[above] (v1) {$\vec{v_1}$};
		% line between v1 and v2
		\draw[] (2,0) -- (1,1) node[above] (phi) {} -- (0,2);
		%Angles
		\draw[black,line width=0.2mm] (\r,0) arc (0:90:\r);
		\draw[blue,line width=0.2mm] (\r,0) arc (0:72:\r);
		\draw[green,line width=0.2mm] (\r,0) arc (0:45:\r);
		\draw[red,line width=0.2mm] (\r,0) arc (0:19:\r);
				
	\end{tikzpicture}
	\hspace{4mm} % dont pass line space here it will split the figures
	\begin{tikzpicture}[scale=0.57]
		\shade[ball color = gray!40, opacity = 0.4] (0,0) circle (2cm);
		% Define radius
		\def\r{2}
														        
		% Bloch vector
		\draw (0,0) node[circle,fill,inner sep=1] (orig) {} -- (1.414, 1.414) node[circle,fill,inner sep=0.7] (b) {};
		\draw (0,0) node[circle,fill,inner sep=1] (orig) {} -- (0.765, 1.847) node[circle,fill,inner sep=0.7] (c) {};
		\draw (0,0) node[circle,fill,inner sep=1] (orig) {} -- (1.847, 0.765) node[circle,fill,inner sep=0.7] (a) {};
				
		\draw (orig) circle (\r);
		\draw[dashed] (orig) ellipse (\r{} and \r/3);
		%draw arc https://tex.stackexchange.com/questions/175016/how-is-arc-defined-in-tikz
		\draw[black,line width=0.2mm] (\r,0) arc (0:90:\r);
		\draw[blue,line width=0.2mm] (\r,0) arc (0:67.5:\r);
		\draw[green,line width=0.2mm] (\r,0) arc (0:45:\r);
		\draw[red,line width=0.2mm] (\r,0) arc (0:22.5:\r);
		% Axes
		% \draw[->] (orig) -- ++(-\r/5,-\r/3) node[below] (x1) {$x_1$};
		\draw[->] (orig) -- ++(\r,0) node[right] (v2) {$\vec{v_2}$};
		\draw[->] (orig) -- ++(0,\r) node[above] (v1) {$\vec{v_1}$};
		%Angles
				
		\pic [draw=red,text=gray,->,angle radius=1.2cm, "\tiny$\theta_1$"] {angle = v2--orig--a};
		\pic [draw=green,text=gray,->,angle radius=1.2cm,"\tiny$\theta_2$" ] {angle = a--orig--b};
		\pic [draw=blue,text=gray,->,angle radius=1.2cm,"\tiny$\theta_3$" ] {angle = b--orig--c};
		\pic [draw=black,text=gray,->,angle radius=1.2cm,"\tiny$\theta_4$" ] {angle = c--orig--v1};
								            
	\end{tikzpicture}
	\caption[Slerp vs Lerp.]{\emph{Slerp} vs. \emph{lerp}. Left side: linear interpolation between $\vec{v_1}$ and $\vec{v_2}$ with $t\in [0.25,0.5,0.75]$. Right side: spherical interpolation. Note that both the length and arc length of the interpolated vectors are equal in \emph{slerp} but unequal in \emph{lerp}.}
	\label{fig:slerp}
	%	\vskip -0.15in
\end{figure}
		
As shown in Fig.~\ref{fig:slerp}, the \emph{slerp} interpolation follows the great circle path on a $d$-dimensional hyper-sphere (with elevation changes) between two points $\bz_i$ and $\bz_j$. This technique has shown promising results in the context of both VAE and GAN generative models, with both uniform and Gaussian priors \cite{SamplingGenerativeNetworks}.

Why \emph{slerp}? Linear interpolation (\emph{lerp}) is the simplest method to traverse the latent space manifold between two known locations. Often it is used to illustrate learned features that capture the semantics of the dataset \cite{NIPS2012_4824}. %For example, when interpolating between two images, one showing a man with black hair and the other a man with blonde hair, linear interpolation in pixel space smoothly transforms the hair color. 
However, \cite{arvanitidis2017latent} noted that linear interpolation in the latent space is often inappropriate since the latent spaces of most generative models are embedded in high dimensional spaces (over $50$ dimensions). In such a space, linear interpolation traverses locations that are extremely unlikely given the prior, whether Gaussian or uniform.

\textbf{Noise concatenation.} Training from a small sample is more susceptible than ever to random perturbations in the data. To increase training robustness, we concatenate noise to the latent vector such that the input to the generator $G_\theta$ is $[\bz_i, \varepsilon],\quad \varepsilon \sim N(0,\sigma I), \,\bz_i \in \{\bz \in Z : ||\bz||_2 =1\})$, see Fig.~\ref{fig:our_method}.
% In effect, this introduces randomness into the training via the batch normalization layers \cite{ioffe2015batch}. 

\textbf{Transductive learning.}
The setting of transductive learning was introduced by \cite{vapnik1998statistical}. Like semi-supervised learning it aims to exploit unlabeled data, using the unlabeled test set to improve the test set classification.

Specifically, the learning task is defined on a given set of $\ell$ training points $$\{(\bx_1,y_1),(\bx_2,y_2),\ldots ,(\bx_\ell,y_\ell)\},\bx_i \in \mathbb{R}^d,y_i\in \{-1,1\} $$ and a sequence of $k$ test vectors $\{\bx_{\ell+1},\ldots,\bx_{\ell+k}\}$. The goal of the learner is to find among an admissible set of binary vectors the one that classifies the test vectors as accurately as possible. We assume that the set of vectors $\{\bx_{\ell+1},\ldots,\bx_{\ell+k}\}$ and their corresponding true labels is an i.i.d sample drawn according to the same unknown distribution $P(\bx)$ and $P(y|\bx)$. Basically, the core idea in transductive learning is to leverage the implicit information in the instances whose output is required ($\bx_{\ell+1},\ldots,\bx_{\ell+k}$), in order to improve their classification.
		
In the same spirit as the classical setting of transductive learning, we may use the unsupervised data to train the generator of the model to reconstruct the test points while optimizing their reconstruction loss (\ref{eq:recon}), see Fig.~\ref{fig:embed}. 
		
\section{Evaluation Methodology}
\label{sec:evalution}

\subsection{Datastes}

We evaluate our method on three standard benchmarks for image classification. The first two are CIFAR-10 and CIFAR-100 \cite{cifar100}, each includes $50,000$ $32 \times 32$ color images, with 10 or 100 classes and 500 or 5000 images per class respectively. The relatively small size of the images allows us to perform an exhaustive ablation study on this dataset as described in Section~\ref{sec:results}. The second dataset is CUB-200 \cite{WahCUB_200_2011}, which includes high-resolution fine-grained images of 200 species of birds, with only 30 images per class. This makes this dataset a more appropriate testbed for a method that addresses the small sample problem.

\subsection{Experimental Protocol}
\label{sec:experiments}

For each benchmark, we defined a small sample task by sub-sampling the original training set of the corresponding dataset. To allow for comparison with other methods, the subset splits were adapted from \cite{cosinesmalldata}. As classification engine we used, unless otherwise noted, \textit{WideResNet-28} \cite{zagoruyko2016wide} for CIFAR-10 and CIFAR-100, and Resnet50 \cite{he2016deep} for CUB-200. Baseline results were obtained by training the corresponding model using the training set with only standard data augmentation. 

When using our method, we augmented the training set using \textit{GLICO}. Specifically, we start by sampling a mini-batch from the training data in each SGD optimization step. Each example $x_i$ in the mini-batch is used for training with probability $0.5$. Otherwise (with probability $0.5$) it is replaced by a new image obtained by sampling the latent space $G([slerp(\bz_i,\bz_j, t),\varepsilon]))$. $\bz_j$ is the latent representation of some example from the same class $c$ as $x_i$, sampled uniformly from the latent codes of all remaining examples in class $c$. The \emph{slerp} interpolation factor is sampled uniformly from the set $[0.1,0.2,0.3,0.4]$.

We compared our results with state-of-the-art methods that are suitable for the small sample domain, using as much as possible public-domain code. Thus we compared sample augmentation with \textit{GLICO} to image augmentation with Cutout \cite{devries2017improved}, Random Erase \cite{randomerasing} and MixMatch \cite{berthelot2019mixmatch}. In each case we repeated the same procedure as described above, replacing the generation of a new image using \textit{GLICO} by an image obtained from the corresponding augmented set of images. We also evaluated the method described in \cite{cosinesmalldata}, which was explicitly designed to handle small sample, using code provided by the authors. 

\subsection{Implementation Details}

\begin{table*}[htbp]
	\caption{Comparison of Top-1 Accuracy (including STE) for CIFAR-100 and CUB-200 using WideResnet-28 and ResNet-50 respectively, with a different number of training examples per class (labeled data only). The methods used for comparison are described in the text below. Best results are marked in bold. For \cite{cosinesmalldata}, $^*$ indicates that the reported results, as obtained in our experiments using code released by the authors, do not match the results reported by the authors which are therefore listed in parentheses. 
		\label{tab: multi-shot-cifar}}
	\centering
	\begin{small}
		\begin{sc}
			\begin{tabular}{l|cl|cccc|c}
				\toprule
				Dataset   & {\footnotesize Samples/Class} & Baseline       & Ours                    & MixMatch & Cutout                  & Random Erase   &              
				\cite{cosinesmalldata}$^*$ \\
				%				\midrule
				%			    \multicolumn{7}{c}{CIFAR-100}                   \\
				\cmidrule(r){1-8}
				CIFAR-100 & 10                            & 22.89$\pm$0.09 & \textbf{28.55$\pm$0.40} & 24.80    & 23.43$\pm$0.24          & 23.26$\pm$0.27 & {23.01 (22)} \\
				          & 25                            & 38.39$\pm$0.10 & \textbf{43.84$\pm$0.25} & 40.17    & 39.11$\pm$0.59          & 37.45$\pm$0.15 & 28.05 (35)   \\
				          & 50                            & 47.82$\pm$0.11 & \textbf{52.95$\pm$0.20} & 49.87    & 52.11$\pm$0.28          & 50.50$\pm$0.41 & 44.55 (48)   \\
				          & 100                           & 61.37$\pm$0.13 & 64.27$\pm$0.04          & 59.03    & \textbf{64.49$\pm$0.10} & 64.03$\pm$0.22 & 55.99 (58)   \\
				%				    \midrule
				%   \multicolumn{7}{c}{CUB-200 } \\
				\cmidrule(r){1-8}
				CUB-200   & 5                             & 50.79$\pm0.19$ & \textbf{51.52$\pm$0.21} & 15.01    & 50.63$\pm$0.31          & 48.90$\pm$0.45 & 17.80 (35)   \\
				          & 10                            & 64.11$\pm0.22$ & \textbf{65.13$\pm$0.12} & 36.02    & 64.33$\pm$0.02          & 63.72$\pm$0.20 & 34.23 (60)   \\
				          & 20                            & 69.11$\pm$0.55 & \textbf{74.16$\pm$0.17} & 60.57    & 68.47$\pm$0.20          & 66.14$\pm$0.23 & 52.00 (76)   \\
				          & 30                            & 75.15$\pm0.10$ & \textbf{77.75$\pm$0.20} & 70.41    & 74.97$\pm$0.34          & 73.74$\pm$0.34 & 62.25 (82)   \\
				\bottomrule
			\end{tabular}
		\end{sc}
	\end{small}
	\vskip -0.15in
\end{table*}
		
% \begin{table*}[htbp]
% 	\caption{Comparison of  Top-1 Accuracy (including STE) for CUB-200 using ResNet-50, with a different number of training examples per class (labeled data only). The methods used for comparison are described in the text below, including some caveats. The best results are marked in bold.
% 		\label{tab: multi-shot-cub}}
% 	\centering
		
% 	\begin{small}
% 		\begin{sc}
% 			\begin{tabular}{cl|cccc|c}
% 				\toprule
																						                 
% 				Samples/Class & Baseline       & Ours                    & MixMatch & Cutout         & Random Erase   & \cite{cosinesmalldata} \\
% 				\midrule
% 				5             & 50.79$\pm0.19$ & \textbf{51.52$\pm$0.21} & 15.01    & 50.63$\pm$0.31 & 48.90$\pm$0.45 & 17.80 (35)           \\
% 				10            & 64.11$\pm0.22$ & \textbf{65.13$\pm$0.12} & 36.02    & 64.33$\pm$0.02 & 63.72$\pm$0.20 & 34.23 (60)           \\
% 				20            & 69.11$\pm$0.55 & \textbf{74.16$\pm$0.17} & 60.57    & 68.47$\pm$0.20 & 66.14$\pm$0.23 & 52.00 (76)           \\
% 				30            & 75.15$\pm0.10$ & \textbf{77.75$\pm$0.20} & 70.41    & 74.97$\pm$0.34 & 73.74$\pm$0.34 & ~62.25~ (82.5)       \\
																						
% 				\bottomrule
% 			\end{tabular}
% 		\end{sc}
% 	\end{small}
	
% \end{table*}

We used SGD optimization with learning rate 0.1 for CIFAR-10 and CIFAR-100 and 0.001 for CUB-200, with a batch size of 128 and 16 respectively. In all the experiments we used the standard categorical cross-entropy loss function when training the weak classifier $f\phi$. (We note in passing that using a strong classifier decreased the quality of the generated images and harmed the final performance, see Section~\ref{sec:weak-vs-strong}.) Images from the CUB-200 dataset were resized to 256 pixels wide in their smaller side, and then randomly cropped to $224\times 224$ pixels. As stated above, in all the experiments (including baseline) we adopted the standard transformations of random horizontal flipping and random crop for data augmentation. 

In all cases, the latent space $Z$ was the unit sphere in ${\mathbb R}^{128}$. For the generator, we used a standard off-the-shelf DCGAN architecture \cite{radford2015unsupervised}. The generator of any modern GAN architecture can be readily used instead and improve the reconstruction quality. However, it is worth noting that our method can improve the SOTA while using the generator of a relatively simple GAN architecture. We trained the model on 2 $\times$ RTX-2080 GPUs for 200 epochs; every epoch took around 40 seconds on CIFAR-100 and 3 minutes on CUB-200.
		
To achieve uniformity we oversampled the training set in proportion to the size of the training set. For example, with 50 samples per class in CIFAR-100, we trained the model $\times 10$ iterations. Standard errors (STE) were obtained from 3 runs with different seeds in all study cases except for MixMatch, where a single result is reported since each MixMatch run took a very long time.

\section{Results}
\label{sec:results}

\begin{figure}[htbp]
% 		\vskip -0.1in
	\centering
	\subfigure[CIFAR-100 image generation.]{

			\includegraphics[width=0.77\linewidth]{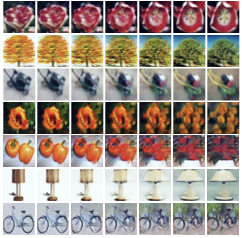}} 
	\qquad
	\subfigure[CUB-200 image reconstruction.]{{
	\includegraphics[width=0.75\linewidth]{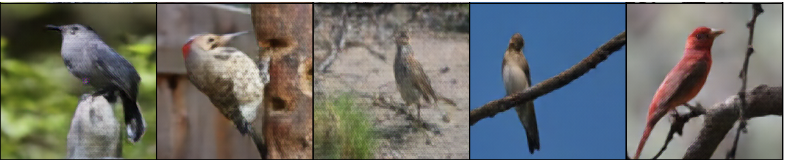}}}

	\caption{Examples of synthesized images. (a) Each row shows five new images (the intermediate columns), generated based on smooth interpolation in the latent space between two reconstructed images (the left and right columns).
		(b) High-resolution image reconstruction. Here \textit{GLICO} was trained only on \textit{10 examples} per class from CUB-200. \label{fig: cub_and_cifar_slerp}}
	% 		\label{fig: cifar_inter}

	\vskip -0.2in
\end{figure}

\begin{figure}[htbp]
	\centering

	\includegraphics[width=0.85\linewidth]{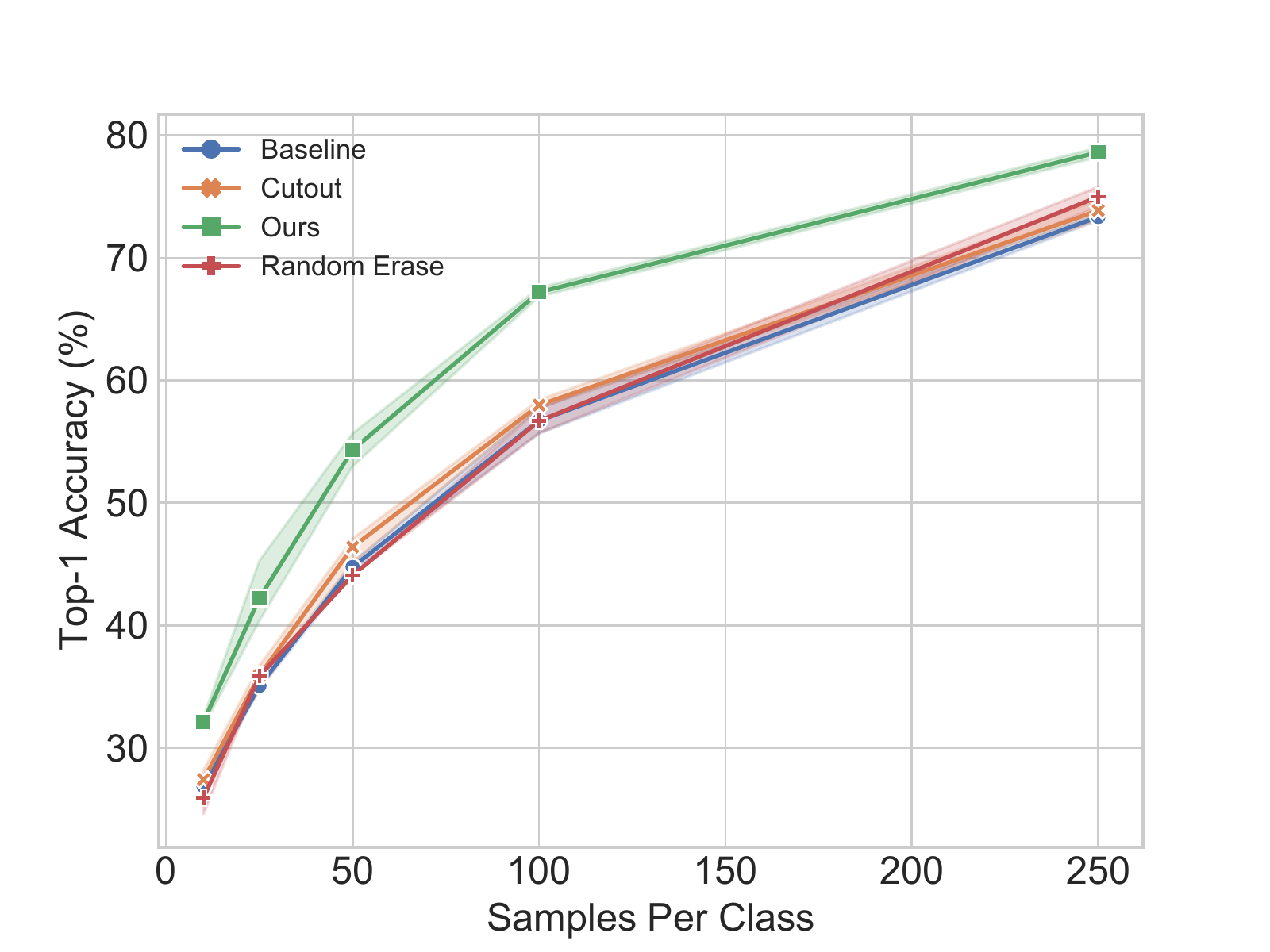}

	\caption{Comparison of  Top-1 Accuracy (including STE) for CIFAR-10 using WideResnet-28, with a different number of training examples per class (labeled data only).}
	\label{fig: cifar10}
	\vskip -0.2in
\end{figure}

The results of our empirical study are summarized in Fig.~\ref{fig: cifar10} for CIFAR-10, and Table~\ref{tab: multi-shot-cifar} for CIFAR-100 and CUB-200. We used small sample partitions, with 10 to 250 labeled samples per class (SPC) in CIFAR-10, 10 to 100 labeled SPC in CIFAR-100, and 5 to 30 labeled SPC in CUB-200. We compared our model to three different methods of data augmentation and one small sample method. 

Fig.~\ref{fig: cub_and_cifar_slerp}a Illustrates the kind of synthetic images we get while relying on a very small sample. Note that since the purpose of generating new images is to boost classification from small sample, low image quality does not preclude their usefulness for the task. Fig.~\ref{fig: cub_and_cifar_slerp}b Illustrates the quality of the reconstruction when using \textit{GLICO} with 10 samples per class only, demonstrating high quality reconstruction. 
		
The two augmentation methods used for comparison in our experiments are Random Erasing \cite{randomerasing} and Cutout \cite{devries2017improved}. In all the study cases but one, augmentation with \textit{GLICO} significantly outperforms the other methods (see Table~\ref{tab: multi-shot-cifar}). With CIFAR-100 and 100 samples per class, Cutout matches the highest accuracy (similar to \textit{GLICO}) in our experiments.

MixMatch \cite{berthelot2019mixmatch} is a new technique that achieves state-of-the-art results on multiple datasets in a semi-supervised setting. In the supervised small sample regime, this method does not perform very well as can be seen in Table~\ref{tab: multi-shot-cifar}, but see Section~\ref{sec:other methods}. Possibly, the blending of images in pixel space, which is intended to provide some means of regularization, is only effective when enough training data are available. Otherwise, it feeds noisy examples to the model and makes it harder to generalize.%It shines when given a large set of unlabeled data and eventually outperforms \textit{GLICO}, see Section~\ref{sec:other methods}. 

\cite{cosinesmalldata} describes a distance-based method that is designed to handle the small sample challenge, among other things. The results, when using the code published by the authors in our experimental design, are shown in the last column of Table~\ref{tab: multi-shot-cifar}. Admittedly, we were not able to reproduce their published results, which are therefore noted in parentheses\footnote{The original published experiments used Resnet-110 for CIFAR-100 rather than the WideResNet-28 model used here, and Resnet-50 for CUB-200 as in our experiments.}.%as estimated from. We also tried to use this paper's main suggestion by using cosine loss \ref{tab:alternatives}. Yet, the results were worse compared to cross-entropy classification loss.

\subsection{Ablation Study}
\label{sec: ablation}

Next, we review and evaluate different design choices used in the architecture and the approach proposed in this paper. The results are summarized in Table~\ref{tab:ablation-study}.

\begin{table}[htbp]
	\centering
	\caption{Ablation Study: Trained on CIFAR-100 with 25 labeled training examples per class. Top 1 and Top 5 accuracy is calculated based on the architectural variants described Section~\ref{sec: ablation}. 	\label{tab:ablation-study}}
	\begin{tabular}{l|ll}
		Model          & Top 1 Acc.     & Top 5 Acc.     \\
		\toprule
		\textit{GLICO} & 43.84$\pm$0.25 & 70.73$\pm$0.07 \\
		Baseline       & 38.39$\pm$0.18 & 67.77$\pm$0.18 \\
		No Classifier  & 41.57$\pm$0.54 & 69.55$\pm$0.11 \\
		No Noise      & 43.31$\pm$0.02 & 70.05$\pm$0.02 \\
		Lerp           & 43.01$\pm$0.06 & 70.51$\pm$0.03 \\
		Transductive   & 44.79$\pm$0.12 & 71.28$\pm$0.09 \\
		\bottomrule
	\end{tabular}
	%	\vskip -0.2in
\end{table}

More specifically, we see in Table~\ref{tab:ablation-study} the effect of omitting different components of \textit{GLICO}, including classifier $f_\phi$, noise concatenation, and replacing \emph{slerp} by vanilla linear interpolation. We note that when omitting classifier $f_\phi$, the reconstruction loss achieves a better score, but the augmentation fails to generate \textit{'good'} examples to improve the classification. The 'Baseline' case shows the results of training without sampling additional images. 'Transductive" shows the added benefit obtained from including the unlabeled test set in the training of generator $G_\theta$.

\subsection{Additional Design Choices}

In this section, we describe a few alternative design choices that proved less effective, as summarized in Table~\ref{tab:alternatives}.

\begin{table}[htbp]
	\centering
	\caption{Top 1 and Top 5 accuracy of additional design choices as explained in the text. 
		\label{tab:alternatives}}
	\begin{tabular}{l|ll}
		Model             & Top 1 Acc.     & Top 5 Acc.     \\
		\toprule
		Latent Classifier & 43.08$\pm$0.06 & 70.22$\pm$0.05 \\
		Hypercube Init    & 44.21$\pm$0.61 & 70.89$\pm$0.46 \\
		ResNet Init       & 42.95$\pm$0.22 & 70.10$\pm$0.13 \\
		%encoded y (cf. to LORD &  &  \\
		Additive Noise    & 41.02$\pm$0.43 & 68.10$\pm$0.11 \\
		%		Noise2noise    & 42.33$\pm$ 0.56  & 70.10$\pm$0.42  \\
		Cosine Loss       & 41.01$\pm$0.27 & 68.45$\pm$0.31 \\
										
		\bottomrule
	\end{tabular}
		
\end{table}

\textbf{Latent Classifier.} One can optimize the discriminative loss $\mathcal{L}_{CE}$ directly in the latent space using $(\bz_i,y_i)$ instead of the image space  $(G(\bz_i),y_i)$. Here we used a 3 layer fully connected network with inter-layer ReLU activation.% between every pair of layers.

\textbf{$Z$ Initialization.} We investigated different ways to initialize the latent space mappings while exploiting some prior knowledge we have on the data. \begin{enumerate*}[label={\roman*)},font={\color{black}\bfseries}]
\item Hypercube vertices: every class is initialized in the vicinity of a different vertex of the hypercube in ${\mathbb R}^{d}$. \item ResNet: each image is assigned the corresponding activation in the penultimate layer of a pre-trained ResNet model.\end{enumerate*}

\textbf{Additive Noise.} \textit{GLICO} relies on the concatenation of noise to the latent space representation. To investigate the contribution of this component, and following \cite{noise2noise}, we explored a simpler alternative, where random noise $\varepsilon \sim \mathcal{N}(0,\sigma I)$ is sampled i.i.d and added to $\bz_i$ before calculating the loss (\ref{eq:recon}), so that $\hat{x}=G(z_i+\varepsilon)$. The goal is to obtain a better representation of the image manifold by learning the $\varepsilon$ ball around every example both in the latent space and the image space. However, as shown in Table~\ref{tab:alternatives}, this approach leads to performance degradation in the final classification. %In \ref{tab:ablation-study} we compared the performances on noise projection before the concatenation to the input for $G$.

\textbf{Cosine Loss.} It is argued in \cite{cosinesmalldata} that the cosine loss provides a better optimization function for the small sample regime. In our experimental setup, the cross-entropy classification loss provided better results, see Table~\ref{tab:alternatives}.

\subsection{Relation to Classical Augmentations}
%\subsection{A non-trivial samples synthesizing}
\label{sec: autoaug}
So far we have shown that our method boosts classification performance in the small sample settings when augmenting the small training set with images synthesized by our generative model. But are we learning to generate any significant new information? In other words, can similar images and the same boost in performance be obtained by simpler alternative means of image augmentation? 

We approach this question by reevaluating the results of our method, modified so that new images are synthesized by an alternative image augmentation technique which employs classical geometric transformation. To make the challenge as hard as possible, we adopt AutoAugment (AA) \cite{Cubuk2018AutoAugmentLA}, an RL based augmentation method, which estimates the optimal set of classical transformations to augment images in CIFAR-100. AA exhaustively searches through 16 types of color-based and geometric base transformations, while using \emph{all the images} in CIFAR-100 benchmark dataset. Note that this gives an unfair advantage to this method as compared to our original method. The case studied here is CIFAR-100 with 50 labeled samples per class, and with transductive learning (similar results are obtained without transductive learning).

\begin{table}[htbp]
	\vskip -0.1in
	\centering
	\caption{Top-1 and Top-5 accuracy when augmenting a small dataset (CIFAR-100, 50 SPC) by \textit{GLICO} alone (second row), AutoAugment alone (third row), or both (fourth row). Note that each method boosts performance on its own, while when used in conjunction additional performance boost is seen. The first row provides the baseline.
	}
	\label{tab: autoaugment}
	\begin{tabular}{ccll}
		\toprule
		AutoAu.      & Ours         & Top-1 Accuracy          & Top-5 Accuracy \\
		\midrule
		             &              & 50.37$\pm$0.05          & 75.61$\pm$0.01 \\
		             & $\checkmark$ & 53.35$\pm$0.23          & 77.60$\pm$0.12 \\
		$\checkmark$ &              & 53.80$\pm$0.10          & 79.18$\pm$0.13 \\
		$\checkmark$ & $\checkmark$ & \textbf{56.31$\pm$0.02} & 80.66$\pm$0.04 \\
		\bottomrule
	\end{tabular}
	\vskip -0.1in
\end{table}
		
The results of this challenge are shown in Table~\ref{tab: autoaugment}. Clearly each method boosts classification performance, as can be seen in rows 2-3 when compared to the baseline in row 1. But when using the two methods - AA and \textit{GLICO} - together, performance improves even further (row 4). It appears that each augmentation method provides an independent contribution, and that the effect of the two augmentation methods is additive. From this empirical result we conclude that the contribution of \textit{GLICO} goes beyond the contribution of augmentation by classical image transformations. %Note that AutoAugment is trained on the entire training set of CIFAR-100 so the direct comparison of our method to this policy is not applicable.

\subsection{Using Unlabeled Data}		
\label{sec:other methods}

While our method is designed to address the strict small sample settings, the same approach can be beneficial in the semi-supervised settings (SSL), where the learner is given access in addition to a large set of unlabeled images.

Note that since \textit{GLICO} is not designed to resolve the SSL problem, it does not have any mechanism of 'label guessing' like other SSL methods. The following analysis simply aims to explore the limits of the current model in different settings. 
Thus, while in the fully supervised scenario \textit{GLICO} outperforms MixMatch as shown in Table~\ref{tab: multi-shot-cifar}, MixMatch slowly improves when given access to unlabeled data, and eventually outperforms \textit{GLICO} as shown in Table~\ref{tab: semisuprvised}. Clearly MixMatch benefits from unlabeled data more considerably. In order to be effective in the semi-supervised settings, we will need to enhance \textit{GLICO} with some mechanism of label guessing. %We note that \textit{GLICO} can benefit from using the test data during training following the transductive learning procedure, which is another way of using unlabeled data to boost training.

\begin{table}[htbp]
	\centering
			
	\caption{Semi-supervised scenario:\MakeLowercase{ Top-1 Accuracy of MixMatch vs GLICO when shown CIFAR-100 with 25 labeled examples per class and a varying number of unlabeled examples, where each case corresponds to a different column.}
	}	
	\label{tab: semisuprvised}
	%	\begin{small}
	\begin{tabular}{l|ccc}
		\toprule
		Method   & supervised only & 1000 unlabeled & 35k unlabeled  \\
		\midrule
		Baseline & 38.39$\pm$0.18  & -              & -              \\
		MixMatch & 40.17           & 42.39          & 50.34          \\
		Ours     & 43.84$\pm$0.25  & 44.52$\pm$0.12 & 44.73$\pm$0.07 \\
																
		\bottomrule
	\end{tabular}
	%	\end{small}
	%	\vskip -0.3in
\end{table}

\subsection{Weak Vs. Strong Classifier}
\label{sec:weak-vs-strong}

Our algorithm is designed to optimize a combined loss $\mathcal{L}_{\textit{percep}}+\mathcal{L}_{\textit{ce}}$, where $\mathcal{L}_{\textit{percep}}$ drives the generator to achieve good reconstruction, while $\mathcal{L}_{\textit{ce}}$ serves as a regularizer. $\mathcal{L}_{\textit{ce}}$ imposes structure on the latent space reflecting the known labels. The minimization of the regularizer $\mathcal{L}_{\textit{ce}}$ is mediated by a classifier $f_\phi$, while the minimization of the reconstruction loss $\mathcal{L}_{\textit{percep}}$ is mediated by a generator $G_\theta$. A strong classifier $f_\phi$, e.g. Resnet-50 or VGG-19 with 45M and 144M parameters respectively, has more than x5 parameters as compared to off-the-shelf generators $G_\theta$. This would shift the balance of the learning algorithm from the generative to the discriminative component of the algorithm. Thus it would seem that the strength of the classifier  $f_\phi$ should be controlled to reflect the amount of labeled data, or how small the sample is. 

\begin{figure}[htbp]
	\vskip -0.2in
	\centerline
{\includegraphics[width=0.85\columnwidth]{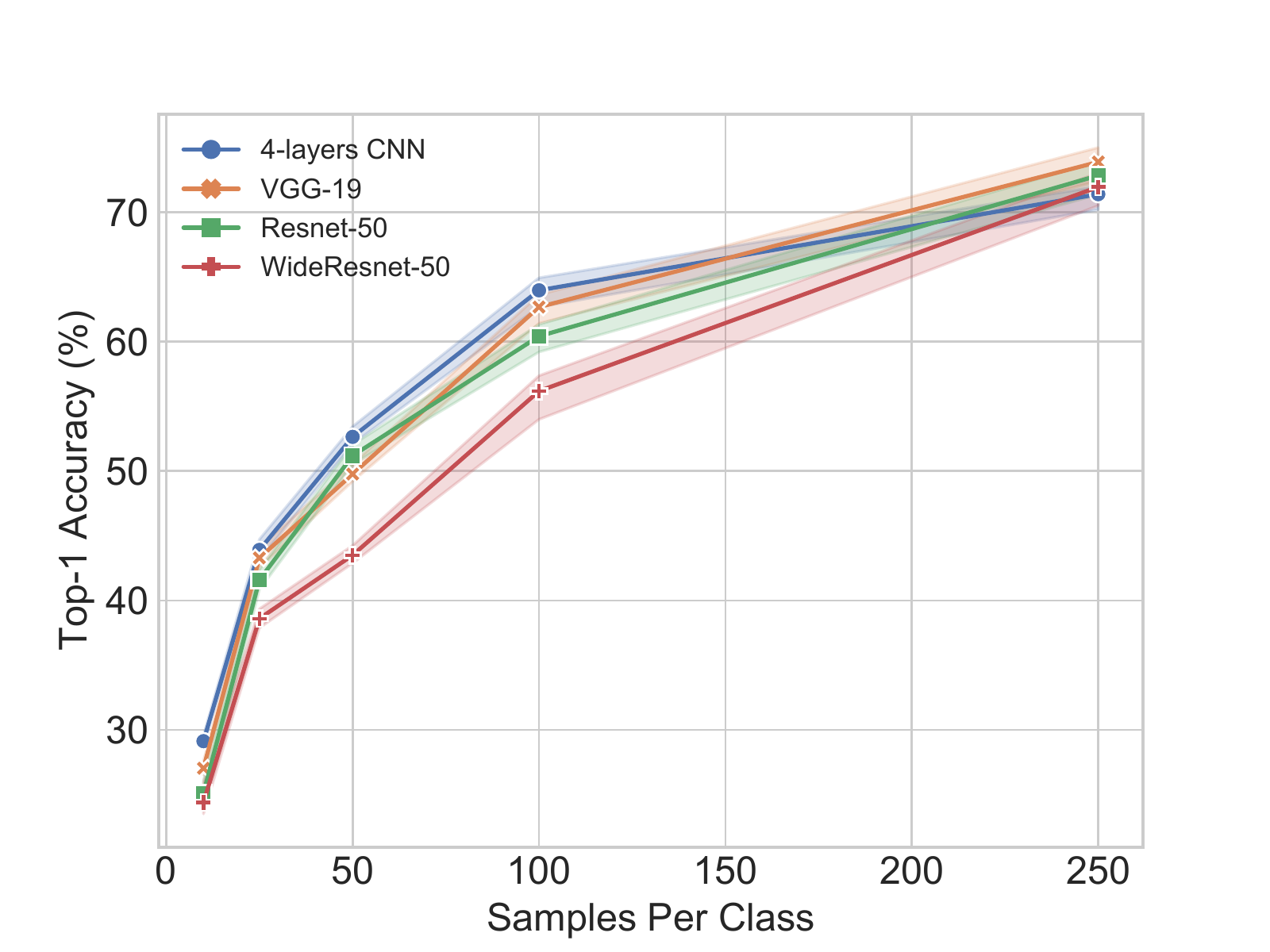}}

	\caption{Weak vs. strong classifier. Top-1 Accuracy (including STE) for  CIFAR-10 when using different architectures for the classifier component in our method. In the small regime (10-100 samples per class) the weak classifier outperforms the stronger classifiers. When sufficient labelled examples are available, the stronger classifiers achieve higher accuracy.   }
	\label{fig: classifiers}
	\vskip -0.1in
\end{figure}

In Fig.~\ref{fig: classifiers} we evaluate four classifiers: 3 strong classifiers including Resnet-50, VGG-19 and Wide-Resnet28, and one weak classifier which is a small CNN with 4 convolutional layers. Our results show that the smaller model achieves the best results  in the low regime of the small sample settings, while VGG19 achieves higher accuracy when more labeled data are available.

\section{Summary and Discussion}

In this work we revisited the problem of learning from small sample. We developed a deep generative model, called Generative latent implicit conditional optimization (\textit{GLICO}), which can be effectively trained to generate examples when seeing only a small sample of data. New examples are synthesized by interpolating between the latent vectors of known examples. When using small sample scenarios generated from the CIFAR-10, CIFAR-100, and CUB-200 benchmarks, we show that our method improves classification over the baseline and several alternative methods. Thus our method defines the state of the art in small sample image classification.

Our generative model is based on latent space optimization. Latent optimization does not involve an encoder like some other generative methods (such as the Variational Auto Encoder). In particular, this implies that the number of variables grows linearly with the number of data samples. Contrary to GAN, latent optimization learns every latent representation separately, and therefore it does not require much data to achieve decent reconstruction results as demonstrated in Fig.~\ref{fig: cub_and_cifar_slerp}. 

The optimization of each representation vector separately also implies that the dimensions of the latent space do not correspond to the semantic features of the data. %, and therefore random sampling of the latent space is not likely to generate useful new images. 
To address this weakness and inject some semantic structure into the latent space representation, we added a classifier to the latent optimization training process. Unlike GANs, the classifier is not trained in an adversarial fashion. Rather, we use the classification loss $\mathcal{L}_{CE}$ over the reconstructed examples $G_\theta(\bz_i)$ to induce semantic relations into the latent space, and allow for better sampling and new image generation (see Fig.~\ref{fig:embed}).

The unique aforementioned properties of our model allow it to improve the training efficacy of deep classifiers in the small sample regime. We suggest two complementary approaches using our proposed transductive learning option. It can be used in conjunction with our method and may benefit from unlabeled data in a semi-supervised manner, or unrelated labeled datasets which can be used for transfer learning. 

\section*{Acknowledgements}
This work was supported by a grant from the Israel Science Foundation (ISF), a grant from the Israel Innovation Authority (Phenomics), and by the Gatsby Charitable Foundations.
%This work was supported in part by a grant from the Israel Science Foundation (ISF) and by the Gatsby Charitable Foundations. 

% \vspace*{-1\baselineskip}
\bibliographystyle{IEEEtran}
\bibliography{IEEEabrv,root}
\vspace{12pt}

\end{document}